\documentclass[conference]{IEEEtran}
\IEEEoverridecommandlockouts

\usepackage{cite}
\usepackage{balance}
\usepackage{hyperref}
\usepackage[numbers]{natbib}
\usepackage{times}
\usepackage{latexsym}
\usepackage{algorithm}
\usepackage[noend]{algpseudocode}
\usepackage[T1]{fontenc}
\usepackage[T5]{fontenc}
\usepackage[utf8]{inputenc}
\usepackage{microtype}
\usepackage{inconsolata}

\usepackage{amsmath}
\usepackage{multirow}
\usepackage{graphicx}
\usepackage{arydshln}
\usepackage{fdsymbol}
\usepackage{ifsym}
\usepackage{tablefootnote}
\usepackage{tikz}
\usepackage{wasysym}
\usepackage{booktabs} 
\usepackage{algorithm}
\usepackage[noend]{algpseudocode}
\usepackage{longtable}
\usepackage{subcaption}
\usepackage{caption}
\usepackage[multiple]{footmisc}
\usepackage{bigfoot}

\DeclareNewFootnote{AAffil}[arabic]
\DeclareNewFootnote{ANote}[fnsymbol]

\def\BibTeX{{\rm B\kern-.05em{\sc i\kern-.025em b}\kern-.08em
T\kern-.1667em\lower.7ex\hbox{E}\kern-.125emX}}

\begin{document}

\title{Link Prediction for Wikipedia Articles as a Natural Language Inference Task}


\author{Chau-Thang~Phan$^{1, 2, \ast,}$\textsuperscript{\textsection}, Quoc-Nam~Nguyen$^{1, 2, \dagger,}$\textsuperscript{\textsection}, and Kiet Van Nguyen$^{1, 2, \star,}$\textsuperscript{\textparagraph}\\
$^{1}$Faculty of Information Science and Engineering, University of Information Technology, Ho Chi Minh City, Vietnam \\
$^{2}$Vietnam National University, Ho Chi Minh City, Vietnam \\
\texttt{\{$^{\ast}$20520929, $^{\dagger}$20520644\}@gm.uit.edu.vn},
\texttt{$^{\star}$kietnv@uit.edu.vn}
}

\maketitle
\begingroup\renewcommand\thefootnote{\textsection}
\footnotetext{Equal contribution}
\endgroup
\begingroup\renewcommand\thefootnote{\textparagraph}
\footnotetext{Corresponding author}
\endgroup
\begin{abstract}
Link prediction task is vital to automatically understanding the structure of large knowledge bases. In this paper, we present our system to solve this task at the Data Science and Advanced Analytics 2023 Competition "Efficient and Effective Link Prediction" (DSAA-2023 Competition) \cite{dsaa-2023-competition} with a corpus containing 948,233 training and 238,265 for public testing. This paper introduces an approach to link prediction in Wikipedia articles by formulating it as a natural language inference (NLI) task. Drawing inspiration from recent advancements in natural language processing and understanding, we cast link prediction as an NLI task, wherein the presence of a link between two articles is treated as a premise, and the task is to determine whether this premise holds based on the information presented in the articles. We implemented our system based on the Sentence Pair Classification for Link Prediction for the Wikipedia Articles task. Our system achieved 0.99996 Macro F1-score and 1.00000 Macro F1-score for the public and private test sets, respectively. Our team UIT-NLP ranked 3rd in performance on the private test set, equal to the scores of the first and second places. Our code\footnote{\url{https://github.com/phanchauthang/dsaa-2023-kaggle/}} is publicly for research purposes.
\end{abstract}


\section{Introduction}\label{main:intro}
Wikipedia\footnote{\url{https://www.wikipedia.org/}}, the world's largest collaborative encyclopedia, has become an invaluable resource for obtaining knowledge on a wide range of topics. With millions of articles covering diverse subjects, Wikipedia offers a vast repository of information that is constantly expanding. However, despite its impressive size, the interlinking between articles within Wikipedia is not always comprehensive, leading to gaps in information connectivity.

Link prediction, a fundamental task in network analysis, aims to predict missing links in a given network based on existing connections. In the context of Wikipedia, link prediction becomes particularly relevant as it can help enhance the navigability of the encyclopedia, improve information accessibility, and promote a deeper understanding of related topics. The DSAA 2023 challenge focuses on the link prediction task applied to Wikipedia articles. Additionally, the focus of the DSAA 2023 challenge \cite{dsaa-2023-competition} is to propose methods for link prediction in network-like data structures, such as network reconstruction and network development, using Wikipedia articles as the primary data source.

Natural Language Inference (NLI) is the task of determining whether a "hypothesis" is true (entailment), false (contradiction), or undetermined (neutral) given a "premise" \cite{storks2019recent}. The link prediction for Wikipedia Articles task is defined as giving a sparsified sub-graph of the Wikipedia network and predicting if a link exists between two Wikipedia pages. In this paper, we leverage the inherent similarities between the Natural Language Inference task and link prediction for Wikipedia articles. Drawing upon this connection, we apply the widely used Sentence Pair Classification (SPC) technique, commonly employed in NLI, to the specific context of the link prediction task.

This paper addresses the link prediction challenge for Wikipedia articles by combining SPC based on NLI and pre-processing techniques. Traditional methods for link prediction in Wikipedia often rely on graph-based algorithms or textual similarity measures \cite{KUMAR2020124289}, which may overlook the nuanced relationships embedded within the text of articles. By integrating sentence pair classification and pre-processing techniques, we aim to capture the semantic and contextual information in Wikipedia articles more effectively, improving link prediction accuracy.

Our contributions are summarized as follows:
\begin{itemize}
    \item Firstly, we adopted efficient data pre-processing techniques to cleanse the comments obtained from Wikipedia. The utilization of these techniques serves to elevate the overall quality of the data and yields significant improvements in the extraction of relevant information before training the model for the Link Prediction task.
    \item Secondly, we leverage the similarities between NLI and link prediction for Wikipedia articles by applying the widely used technique of SPC from NLI to link prediction.
    \item Finally, we achieved the best result on this task, accounting for 0.99996 on the public test and 1.00000 on the private test, ranking 8th and 3rd, respectively.
\end{itemize}
\section{Related Works}
\subsection{Link Prediction}
Link prediction, a critical issue in network-structured data \cite{KUMAR2020124289}, was initially introduced as a supervised learning task by \citet{al2006link}. They also identified key performance features within the supervised learning framework. Building upon the concept of analyzing user-item interactions as graphs, \citet{huang2005link} utilized link prediction techniques from the recent network modeling literature to enhance collaborative filtering recommendations. In contrast, \citet{liu2010link} proposed a method based on local random walk, which achieves competitive or even superior predictions compared to other random-walk-based approaches while maintaining lower computational complexity. Another approach suggested by \citet{pmlr-v48-trouillon16} involves solving the link prediction task through latent factorization, enabling scalability to large datasets with linear space and time complexity. Additionally, the heuristic learning paradigm for link prediction was explored by \citet{NEURIPS2018_53f0d7c5}. Furthermore, \citet{Negietal} framed the link prediction task in heterogeneous networks as a multi-task metric learning task.

\subsection{Natural Language Inference}

In the field of artificial intelligence, inference has always been a prominent topic. While significant advancements have been made in automatic methods for formal deduction, the Natural Language Inference (NLI) task has seen comparatively slower progress. NLI refers to the task of determining the validity of inferring a natural language hypothesis ($h$) from a natural language premise ($p$) \cite{maccartney2009natural}. \citet{bowman-etal-2015-large} introduced the Stanford Natural Language Inference corpus, which consists of labeled sentence pairs created by humans engaged in a novel task based on image captioning. This corpus is freely accessible and serves as a valuable resource. \citet{wang-jiang-2016-learning} proposed a specialized architecture based on long short-term memory (LSTM) networks for NLI. Their approach utilizes a match-LSTM that conducts word-by-word comparisons between the hypothesis and the premise, which enhances the performance of NLI. \citet{Chen-Qian-2017-ACL} demonstrated that carefully designing sequential inference models using chain LSTMs can surpass previous models in terms of performance. 
\section{Algorithmic Techniques}
A brief of link prediction for Wikipedia articles task definition, dataset, our pre-processing techniques, and our proposed approach for the DSAA 2023 challenge \cite{dsaa-2023-competition} are presented in this section.
\subsection{Task Definition}
The task involves predicting the presence or absence of an edge between a pair of nodes $(u, v)$. In this competition, our focus is on link prediction for Wikipedia articles. To be more precise, given a sparsified subgraph of the Wikipedia network, the objective is to determine whether a link exists between two pages, $u$ and $v$, in the context of Wikipedia. 
\subsection{DSAA-2023 Dataset}
The dataset used in the contest is extracted from Wikipedia, where graph nodes are annotated with text. The data of the DSAA-2023 Competition include the following:
\begin{itemize}
    \item train.csv file: It contains pairs of nodes along with an indication of whether there is an edge between them. The file has four columns: $id$, $id1$, $id2$, and $label$. The id column represents the pairing identifier, $id1$, and $id2$ are the node IDs, and the label column declares the presence or absence of an edge (0 or 1).
    \item nodes.zip file: This file is a compressed version of nodes.tsv and contains information about each node. It consists of two columns: id (node identifier) and text (textual description of the node).
    \item sample\_submission.csv file: This demo submission file has two columns: the pair ID and the corresponding label ($0$ or $1$).
\end{itemize}

The statistics of the training set labels after combining train.csv and nodes.tsv files together are presented in Table~\ref{fig:labels}.

\begin{table}[ht]
\centering
\begin{tabular}{l|l|l} 
\hline
           & \textbf{Non-related (0)} & \textbf{Related (1)}  \\ 
\hline
Frequency  & 512,389                  & 435,843               \\
Percentage & 54.03 (\%)               & 45.97 (\%)            \\
\hline
\end{tabular}
\caption{Statistics of the Training Set Labels.}
\label{fig:labels}
\end{table}
\subsection{Pre-Processing Techniques}
The dataset consists of Wikipedia objects' main content and CSS codes within Wikipedia. However, these CSS codes are considered noise and are not required for training. Consequently, it is crucial to remove them. To accomplish this, we employ two algorithms, the Balance Curly Bracing and Remove Double Curly Bracing Algorithms, Algorithm~\ref{balance} and Algorithm~\ref{removal}, respectively, to effectively eliminate these unnecessary codes. Punctuation marks, such as periods, commas, question marks, exclamation marks, and others, and redundant spaces can introduce noise and disrupt the natural flow of text during training. Therefore, we apply the Regex library\footnote{\url{https://docs.python.org/3/library/re.html}} to remove all of them. Our preprocessing techniques for this task of the DSAA-2023 Competition are summarized as follows: Balancing curly braces, removing double curly braces, removing all punctations, and removing redundant spaces.


\begin{algorithm}[ht]
\caption{Balancing Curly Braces}
\begin{algorithmic}[1]
\Procedure{BalanceCurlyBraces}{\textit{text}}
    \State $opening\_count \gets$ $text$.count(`\{') 
    \State $closing\_count \gets$ $text$.count(`\}')

    \If{$opening\_count > closing\_count$}
        \While{$opening\_count > closing\_count$}
            \State $index \gets$ $text$.find(`\{')
            \If{$index \neq -1$}
                \State $text$ = $text$[:index] + $text$[index + 1:]
                \State $opening\_count \gets opening\_count - 1$
            \EndIf
        \EndWhile
    \ElsIf{$closing\_count > opening\_count$}
        \While{$closing\_count > opening\_count$}
            \State $index \gets$ $text$.rfind(`\}')
            \If{$index \neq -1$}
                \State $text$ = $text$[:index] + $text$[index + 1:]
                \State $closing\_count \gets closing\_count - 1$
            \EndIf
        \EndWhile
    \EndIf

    \State \textbf{return} $text$.strip()
\EndProcedure
\end{algorithmic}
\label{balance}
\end{algorithm}


\begin{algorithm}[ht]
\caption{Removing Double Curly Braces}
\begin{algorithmic}[1]
\Procedure{RemoveDoubleCurlyBraces}{text}
    \State $stack \gets []$
    \State $clean\_text \gets \text{` '}$
    
    \For{$char$ \textbf{in} $text$}
        \If{$char = \text{`}\{\text{'}$}
            \State $\text{$stack$.push}(char)$
        \ElsIf{$char = \text{`}\}\text{'}$}
            \If{$\text{not isEmpty}(stack)$ \textbf{and}           
                \State $\text{top}(stack) = \text{`}\{\text{'}$} \State $\text{$stack$.pop}(stack)$
            \EndIf
        \Else
            \If{$\text{isEmpty}(stack)$}
                \State $clean\_text \gets clean\_text + char$
            \EndIf
        \EndIf
    \EndFor
    
    \State \textbf{return} $clean\_text$
\EndProcedure
\end{algorithmic}
\label{removal}
\end{algorithm}

\subsection{Our Proposed Approach}
Our approach draws inspiration from recent successes in natural language understanding tasks, particularly in pre-trained language models. We posit that a link between two Wikipedia articles can be treated as a premise, and predicting the link's existence is akin to determining the logical relationship between the premise and a hypothesis. We encode the content of the two articles as the premise and the hypothesis, respectively, and utilize this NLI setup for link prediction.

In sentence pair classification, the model takes two sentences as input and aims to determine their relationship. It learns to classify the pairs into different categories based on the desired task. Link prediction for Wikipedia Articles task is to predict whether there is an edge or link between the two given nodes. 

\begin{figure*}[ht]
    \centering
    \includegraphics[width=0.56\textwidth]{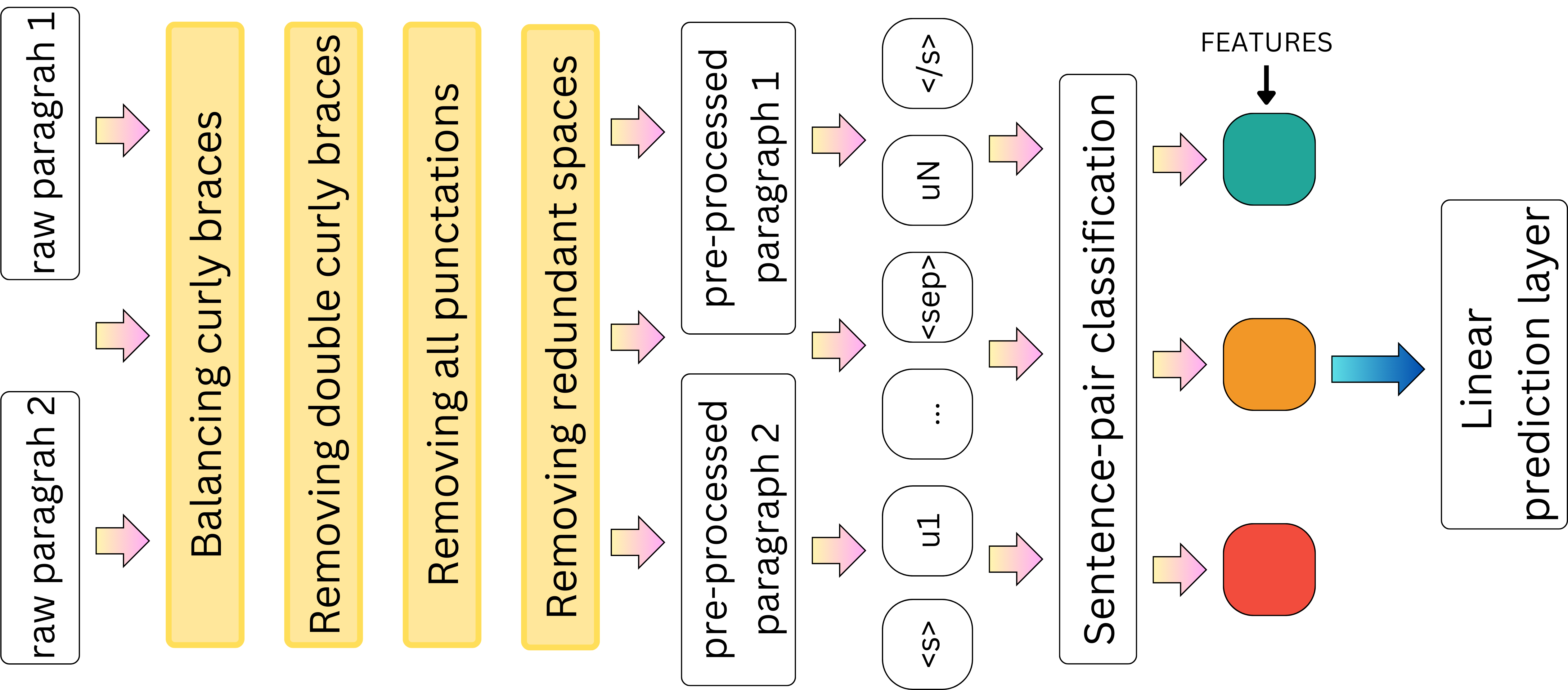}
    \caption{Our Approach Architecture for Link Prediction for Wikipedia Articles.}
    \label{fig:proposed}
\end{figure*}

In this paper, we have implemented a Sentence Pair classification model based on XLM-Roberta, a well-known architecture introduced in the work of Conneau et al. (2020) \cite{conneau-etal-2020-unsupervised}, which is widely used for Natural Language Inference (NLI). Our implementation is tailored for generic and specific Wikipedia article link prediction tasks. Additionally, we incorporate a linear prediction layer on top of the XLM-Roberta architecture to calculate the final output. The details of our proposed approach's architecture are thoroughly presented in Figure~\ref{fig:proposed}.

\section{Experimental Results}
\subsection{Experimental Configurations}
We followed fairly standard practices for fine-tuning, most of which are described in \cite{bert}. We use a batch size of 128, a maximum token length of 128, a learning rate of 2e-5, and AdamW optimizer with an epsilon of 1e-8.

We empirically SentencePair classification model for our system using \textit{simpletransformers}\footnote{\url{https://simpletransformers.ai/} (ver 0.63.11)}. We trained our model on 1$\times$RTX4090 GPU on the Vast.ai\footnote{\url{https://vast.ai}} platform for four hours. Ten seconds is needed for predicting the test set of this dataset.
\subsection{Experimental Results}
In this section, we describe our experiments and results for
Link Prediction task. Only the macro F1-score is used for evaluation. Experimental results are presented in Table~\ref{tab:results}

By implementing robust and efficient data preprocessing techniques tailored explicitly for cleansing comments obtained from Wikipedia, we aimed to enhance the overall data quality and substantially improve extracting relevant information before training the model for link prediction tasks. Combining these pre-processing techniques with the sentence pair classification model resulted in a notable performance improvement compared to scenarios where such techniques were not applied.

Our proposed approach achieved remarkable results, with a macro F1-score of 0.99996 for the public test and a perfect score of 1.00000 for the private test. These exceptional scores indicate the effectiveness of our methodology in accurately predicting links within the given context. In terms of ranking, we secured the 8th position on the public test and an impressive 3rd position on the private test.
\begin{table}[ht]
\centering
\resizebox{!}{0.12\columnwidth}{
\begin{tabular}{l|c|c} 
\hline
            & \textbf{Public test} & \textbf{Private test}  \\ 
\hline
\multicolumn{3}{c}{With Preprocessing techniques}           \\ 
\hline
Our approach & \textbf{0.99996}     & \textbf{1.00000}       \\ 
\hline
\multicolumn{3}{c}{Without Preprocessing techniques}        \\ 
\hline
Our approach & 0.97680              & 0.97663                \\
\hline
\end{tabular}}
\caption{Experimental Results in terms of Macro F1-score.}
\label{tab:results}
\end{table}
\section{Conclusion}

The significance of the link prediction task lies in its ability to comprehend the structure of extensive knowledge bases automatically. In this paper, we provide a detailed account of our proposed approach to addressing this task within the context of the Data Science and Advanced Analytics 2023 Competition titled "Efficient and Effective Link Prediction". To tackle the link prediction task for Wikipedia articles, we implemented a sentence pair classification based on natural language inference and an efficient pre-processing techniques pipeline. Notably, our system achieved exceptional performance with a macro F1-score of 0.99996 for the public test set and a perfect score of 1.00000 for the private test set. As a result, our system secured ranked 3rd on the private test set, equal to the scores of the first and second places.


\section*{Acknowledgement}
This research was supported by The VNUHCM-University of Information Technology’s Scientific Research Support Fund.

\bibliographystyle{IEEEtranN}
\bibliography{custom.bib}
\end{document}